\documentclass[11pt]{article}
\usepackage[textwidth=400pt,centering]{geometry}
\usepackage{fullpage}
\usepackage{microtype}
\usepackage{graphicx}
\usepackage{subfigure}
\usepackage{booktabs} 
\usepackage{xcolor}
\usepackage[square, sort, numbers]{natbib}
\usepackage[nottoc,notlot,notlof]{tocbibind}
\usepackage{amsmath,amssymb,amsthm}
\usepackage{pifont}
\usepackage{bbm}
\usepackage{multirow}
\usepackage{lmodern}
\DeclareMathOperator*{\argmax}{arg\,max}
\DeclareMathOperator*{\argmin}{arg\,min}
\theoremstyle{definition}
\newtheorem{definition}{Definition}
\newtheorem{theorem}{Theorem}
\newtheorem{lemma}{Lemma}
\definecolor{darkblue}{rgb}{0,0.08,0.8}
\usepackage[utf8]{inputenc} 
\usepackage[T1]{fontenc}    
\usepackage[colorlinks=true,linkcolor=blue, citecolor=blue]{hyperref}
\usepackage{url}            
\usepackage{booktabs}       
\usepackage{amsfonts}       
\usepackage{nicefrac}       
\usepackage{microtype}      

\usepackage{times}  
\usepackage{helvet} 
\usepackage{courier}
\usepackage{graphicx}
\usepackage{datatool}

\usepackage{amsmath,amsthm,amssymb}
\usepackage{pifont}
\usepackage{bbm}
\usepackage{multirow}
\usepackage{enumitem}
 \usepackage{float}
\theoremstyle{definition}
\usepackage{amsmath,amsthm,amssymb}
\usepackage{mathtools}
\usepackage{pifont}
\usepackage{bbm}
\usepackage{multirow}
\usepackage{enumitem}
\usepackage[linesnumbered,ruled,vlined]{algorithm2e}
 \usepackage{float}
\newcommand\numberthis{\addtocounter{equation}{1}\tag{\theequation}}

\begin{document}
\title{Domain Discrepancy Measure for Complex Models \\in Unsupervised Domain Adaptation}
\author{Jongyeong Lee$^{1,2}$ \and Nontawat Charoenphakdee$^{1,2}$  \and Seiichi Kuroki$^{1,2}$ \and Masashi Sugiyama$^{2,1}$}
\date{
$^1$ The University of Tokyo 
$^2$ RIKEN
}
\maketitle

\begin{abstract}
Appropriately evaluating the discrepancy between domains is essential for the success of unsupervised domain adaptation. In this paper, we first point out that existing discrepancy measures are less informative when complex models such as deep neural networks are used, in addition to the facts that they can be computationally highly demanding and their range of applications is limited only to binary classification. We then propose a novel domain discrepancy measure, called the \emph{paired hypotheses discrepancy} (PHD), to overcome these shortcomings. PHD is computationally efficient and applicable to multi-class classification. Through  generalization error bound analysis, we theoretically show that PHD is effective even for complex models. Finally, we demonstrate the practical usefulness of PHD through experiments.
\end{abstract}
\section{Introduction}
Deep learning has demonstrated its flexibility and effectiveness in real world applications such as speech recognition~\citep{hinton2012deep}, computer vision~\citep{krizhevsky2012imagenet} and game playing~\citep{silver2016mastering}.
This success is partially attributed to label-rich datasets and based on the assumption that the source and target domains are identical.
However, in many challenging applications in natural language processing~\citep{jiang2007instance}, speech recognition~\citep{deng2014autoencoder}, and computer vision~\citep{bousmalis2017unsupervised}, labeling data is highly expensive and the two domains are often quite different. In particular, when the source and target domains are different from each other, learning from source data might lead to performance degradation in the target domain.
To deal with this scenario, \emph{unsupervised domain adaptation} has been extensively studied~\citep{pmlr-v28-germain13, pmlr-v37-ganin15,kodirov2015unsupervised,long2016unsupervised}. 
\par
Generally, domain adaptation shows a good performance when the source and target domains are similar~\citep{pan2010survey}. 
Therefore, an important topic for domain adaptation is how to measure the difference between two domains. 
So far, many discrepancy measures have been proposed~\citep{huang2007correcting, sugiyama2008direct, mansour2009multiple, zhang2012generalization, courty2017joint, cortes2019adaptation}.
\par
In unsupervised domain adaptation, we cannot directly measure domain discrepancy using labeled data in the target domain since we have only unlabeled data in the target domain. 
One way is to measure the difference between two domains without using any label information.
First, \citet{ben2007analysis} proposed a discrepancy measure which explicitly considers hypothesis classes in a binary setting with the zero-one loss. 
Following this research, \citet{mansour2009domain} generalized the discrepancy measure of \citet{ben2007analysis} to arbitrary loss functions.
These discrepancy measures which explicitly consider hypothesis classes give tighter  generalization error bounds than others which do not use information of the hypothesis class such as the $L_1$ distance and Wasserstein distance~\citep{arjovsky2017wasserstein, AAAI1817155}. 
However, the discrepancy measure of \citet{mansour2009domain} requires high computation costs, while the discrepancy measure by \citet{ben2007analysis} has no theoretical guarantee.
\par
To alleviate these problems, \citet{kuroki2018unsupervised} proposed a computationally efficient discrepancy measure which utilizes label information in the source domain. 
Nevertheless, we show that existing discrepancy measures mentioned above may induce loose  generalization error bounds and their estimation can be unreliable when complex models such as deep neural networks are applied.
\par
In this paper, to overcome the limitations of the existing discrepancy measures, we propose a novel discrepancy measure named \emph{paired hypotheses discrepancy} (PHD), which only considers a fixed pair of hypotheses, not the whole hypothesis class. 
By incorporating unlabeled target data with labeled source data to make a reliable hypothesis, PHD is informative even for complex models such as multilayer perceptrons (MLPs) and convolutional neural networks (CNNs). 
\par
Our contributions are summarized as follows: 
\begin{itemize}[wide, labelindent = 3pt, topsep = 1ex]
    \item We propose a novel discrepancy measure, PHD, which considers a pair of hypotheses and can be effectively applied to complex models.
    \item We show that PHD can be applied to any loss functions satisfying the triangle inequality, including the zero-one loss, and it can handle multi-class problems.
    \item We derive generalization error bounds in the target domain for PHD.
    \item We demonstrate the effectiveness of PHD for neural networks in experiments.
\end{itemize}

\section{Preliminaries}
In this section, we formulate the problem and review existing methods.

\subsection{Problem setting and notation}
Here, we introduce the notations used to formulate the problem of unsupervised domain adaptation.
Let $\mathcal{X}$ be the input space and $\mathcal{Y}$ be the output space, which is $\{+1, -1\}$ in binary classification and $\{1, \ldots, k\}$ in $k$-class classification.
We also define a domain as pair ($P_\mathrm{X}, f_\mathrm{X}$), where $P_\mathrm{X}$ is the input distribution on $\mathcal{X}$ and $f_\mathrm{X} : \mathcal{X} \rightarrow \mathcal{Y}$ is a true labeling function. In unsupervised domain adaptation, 
we have the following data:
\begin{itemize}[wide, labelindent = 3pt, topsep = 1ex]
    \item Labeled source data $\mathcal{S}=\{(x_i^\mathrm{S},f_\mathrm{S}(x_i^\mathrm{S}))\}_{i=1}^{n_\mathrm{S}}$ where $x_i^\mathrm{S}\stackrel{\mathrm{i.i.d.}}{\sim} P_\mathrm{S}$.
    \item Unlabeled target data $\mathcal{T}=\{x_j^\mathrm{T}\}_{j=1}^{n_\mathrm{T}}$ where $x_j^\mathrm{T}\stackrel{\mathrm{i.i.d.}}{\sim} P_\mathrm{T}$.
\end{itemize}
We denote a loss function as $\ell : \mathbb{R} \times \mathbb{R} \rightarrow \mathbb{R}_{\geq 0}$, and the expected loss as $R_{P_\mathrm{X}}^\ell (h,h^{\prime}) = \mathbb{E}_{x\sim P_\mathrm{X}}[\ell(h(x),h^{\prime}(x))]$ for distribution $P_\mathrm{X}$ and labeling functions $h,h^\prime$ in hypothesis class $\mathcal{H} : \mathcal{X} \rightarrow \mathbb{R}$.
Also we denote the empirical loss as $\widehat{R}^\ell_\mathrm{X} (h,h^{\prime}) = \frac{1}{\mathrm{|}X\mathrm{|}}\sum\limits_{x \in \mathrm{X}}\ell(h(x),h^{\prime}(x))$ where $X$ is independently sampled from distribution $P_\mathrm{X}$.
We denote the true risk minimizer in the domain ($P_\mathrm{X},f_\mathrm{X}$) in a hypothesis class $\mathcal{H}$ as $h_\mathrm{X}^* = \argmin\limits_{h\in\mathcal{H}}R_{P_\mathrm{X}}^\ell(h,f_\mathrm{X})$.
Note that $h_\mathrm{X}^*$ is not necessarily the same as true labeling function $f_\mathrm{X}$ since the hypothesis class is restricted to $\mathcal{H}$.
The empirical risk minimizer in the domain ($P_\mathrm{X},f_\mathrm{X}$) in a hypothesis class $\mathcal{H}$ is denoted as $\hat{h}_\mathrm{X}=\argmin\limits_{h\in\mathrm{H}}\widehat{R}^\ell_{\mathrm{X}}(h,f_\mathrm{X}).$

\subsection{Existing discrepancy measures}
In unsupervised domain adaptation, it is essential to measure the difference of distributions since the performance of unsupervised domain adaptation might be degraded if the distributions in two domains are significantly different from each other \citep{pan2010survey}. 
However, it may not be impossible to measure the distance between two distributions using both the input space and the output space since labels of target data  (output space) are unknown. 
\par
As a way to deal with this problem, using only input space to measure the distance has been considered.  
\citet{mansour2009domain} proposed a discrepancy measure called the discrepancy distance which explicitly takes the hypothesis class into account. The discrepancy distance  is defined as
\begin{equation*}
  \mathrm{disc}_\mathcal{H}^\ell(P_\mathrm{T}, P_\mathrm{S}) = \sup\limits_{h,h^{\prime} \in \mathcal{H}}\mathcal{|}R_{P_\mathrm{T}}^{\ell}(h,h^{\prime})-R_{P_\mathrm{S}}^{\ell}(h,h^{\prime})\mathcal{|}.  
\end{equation*}
An intuition behind the discrepancy distance is that if the input distributions in the two domains are the same, the behavior of any pairs in each domain will be also same which leads to $\mathrm{disc}=0$.
\par
Also, \citet{mansour2009domain} showed that the following inequalities hold for any $h,h^{\prime} \in \mathcal{H}:$
\begin{align*}
    \mathcal{|}R_{P_\mathrm{T}}^{\ell}(h,h^{\prime})-R_{P_\mathrm{S}}^{\ell}(h,h^{\prime})\mathcal{|} &\leq  \mathrm{disc}_\mathcal{H}^\ell(P_\mathrm{T}, P_\mathrm{S}) \\ &\leq M \cdot L_1(P_\mathrm{T},P_\mathrm{S}),
\end{align*}
where $M$ is a positive constant and $L_1(\cdot, \cdot)$ is the $L_1$-distance over distributions. \citet{redko2017theoretical} showed that the following inequalities hold for every pair of hypotheses $h, h^{\prime}$ in a reproducing kernel Hilbert space $\mathcal{H}$:
\begin{align*}
    \mathcal{|}R_{P_\mathrm{T}}^{\ell}(h,h^{\prime})-R_{P_\mathrm{S}}^{\ell}(h,h^{\prime})\mathcal{|} &\leq  \mathrm{disc}_\mathcal{H}^\ell(P_\mathrm{T}, P_\mathrm{S}) \\ &\leq W_1(P_\mathrm{T},P_\mathrm{S}),
\end{align*}
where $W_1(\cdot,\cdot)$ is the Wasserstein-1 distance over distributions. 
These inequalities imply the suitability of the discrepancy distance for measuring domain discrepancy since this implies the tightness of the generalization error bound.
However, \citet{mansour2009domain} only provided an algorithm for only \emph{1-dimensional data} since computation of the discrepancy distance is time-consuming because it considers the worst pair of hypotheses.
\par
\citet{ben2007analysis} provided a discrepancy measure $d_{\mathcal{H}}$, which is a computationally efficient proxy of the discrepancy distance for the zero-one loss in a binary setting. $d_\mathcal{H}$ is defined as
\begin{equation*}
    d_\mathcal{H}(P_\mathrm{T}, P_\mathrm{S}) = \sup\limits_{h \in \mathcal{H}}\mathcal{|}R_{P_\mathrm{T}}^{\ell_{01}}(h,1)-R_{P_\mathrm{S}}^{\ell_{01}}(h,1)\mathcal{|}.
\end{equation*}
However, differently from the discrepancy distance, $d_\mathcal{H}$ does not provide any learning guarantee.
\par
To alleviate this problem, \citet{kuroki2018unsupervised} proposed a source-guided discrepancy ($\mathrm{S}$-disc) defined as
\begin{equation*}
    S_\mathcal{H}^\ell(P_\mathrm{T}, P_\mathrm{S}) = \sup\limits_{h \in \mathcal{H}}\mathcal{|}R_{P_\mathrm{T}}^{\ell}(h,h_\mathrm{S}^*)-R_{P_\mathrm{S}}^{\ell}(h,h_\mathrm{S}^*)\mathcal{|}.
\end{equation*}
$\mathrm{S}$-disc utilizes the information of label-rich source data to reduce the computation costs with a tighter generalization error bound in the target domain than other existing discrepancy measures. 
Similarly to $d_\mathcal{H}$, however, \citet{kuroki2018unsupervised} only provided estimation of $\mathrm{S}$-disc for the zero-one loss in a \emph{binary setting} which is a critical limitation for practical use.
\par
Moreover, the estimation of discrepancy measures that contains a supremum term may fail miserably when using complex models. Table~\ref{tab:exp_cmp} illustrates the failure of the estimation of existing discrepancy measures.
\par
Here, we performed the same experiments but only the hypothesis class
(model) was changed, i.e., the linear model (Linear) or a multi-layer perceptron (MLP).
Although the discrepancies are supposed to be small since two domains are exactly identical, $\mathrm{S}$-disc and $d_\mathcal{H}$ highly overestimate the discrepancy when the hypothesis class is an MLP, even though they outputted small values for the linear model. 
This shows that small difference between the input distributions might be overestimated when complex models are applied to the existing methods.
This suggests that the instance-reweighting method based on existing discrepancy measures may mistakenly adapt instances in the same domain as if they are from the different domains when employing a complex model.
On the other hand, our proposed discrepancy measure which is introduced in Section~\ref{sec:proposed} does not suffer from this problem.

\begin{table}[t]
\caption{Failure of existing discrepancy measures when the source and target domains are identical. MNIST~\citep{lecun1998mnist}  and EMNIST~\citep{cohen2017emnist} were used. Note that the empirical discrepancy measures should be close to zero. Here, $h_\mathrm{S}$ indicates the empirical accuracy of the source empirical risk minimizer. PHD is our proposed discrepancy measure. The mean and standard deviation of 10 trials are reported.}
\begin{small}
\label{tab:exp_cmp}
\begin{center}
\begin{tabular}{cccccc}
     \toprule
     &&&\multicolumn{3}{c}{Discrepancy measure}\\
       Dataset & Model  & $h_\mathrm{S}$(\%) & $d_\mathcal{H}$ & $\mathrm{S}$-{disc} & $\mathrm{PHD}_\mathrm{PNU}$  \\
    \midrule
     \multirow{4}{*}{MNIST} & \multirow{2}{*}{Linear} & 90.36& 0.1169  & 0.0056  & 0.0147 \\
     && (0.139) & (0.004) & (0.002) & (0.008) \\
      & \multirow{2}{*}{MLP}& 97.859  & 0.8783 & 0.8449 & 0.0128 \\
      && (0.054)& (0.014) & (0.015) & (0.008)\\
    \midrule
         \multirow{4}{*}{EMNIST} &\multirow{2}{*}{Linear} & 89.85 & 0.0353 & 0.0023  & 0.0131\\
         &&(0.388) & (0.003) & (0.001) & (0.007)\\
      & \multirow{2}{*}{MLP} & 99.1 & 0.3767 & 0.3696 & 0.0060\\
      & & (0.017) & (0.018) & (0.014) & (0.001) \\
      \bottomrule
\end{tabular}
\end{center}
\end{small}
\end{table}

\section{Generalization Error Bounds of the Existing Methods}\label{sec3}
This section provides a closer look at  generalization error bounds based on the existing discrepancy measures since the goal of unsupervised domain adaptation is to minimize the  generalization error bound in the target domain without any target label information. 
It has been suggested in the literature that taking a hypothesis class into account can make the bound tighter \citep{mansour2009domain, kuroki2018unsupervised}. Nevertheless, we point out that the existing discrepancy measures may cause the bound to be less informative when applying a complex model.

To relate the source domain to the target domain, the existing  generalization error bounds utilize the loss $\ell$ that satisfies the triangle inequality, e.g., the zero-one loss for classification and the $\ell_1$ loss in regression. 
If a loss $\ell$ satisfies the triangle inequality, the following  generalization error bound can be obtained by applying the triangle inequality:
\begin{equation*}
    R_{P_\mathrm{T}}^\ell(h, f_\mathrm{T}) - R_{P_\mathrm{T}}^\ell(h_\mathrm{T}^*, f_\mathrm{T}) \leq R_{P_\mathrm{T}}^\ell(h,h_\mathrm{T}^*) \leq 
    R_{P_\mathrm{T}}^\ell(h,h_\mathrm{S}^*) + R_{P_\mathrm{T}}^\ell(h_\mathrm{S}^*, h_\mathrm{T}^*). \numberthis \label{bound}
\end{equation*}
Upper-bounding the RHS of (\ref{bound}) as follows leads to S-disc  \citep{kuroki2018unsupervised}:
\begin{equation}\label{bound_2}
    R_{P_\mathrm{T}}^\ell(h,h_\mathrm{S}^*) + R_{P_\mathrm{T}}^\ell(h_\mathrm{S}^*,h_\mathrm{T}^*) + R_{P_\mathrm{S}}^\ell(h,h_\mathrm{S}^*) - R_{P_\mathrm{S}}^\ell(h,h_\mathrm{S}^*) \leq R_{P_\mathrm{S}}^\ell(h,h_\mathrm{S}^*) + R_{P_\mathrm{T}}^\ell(h_\mathrm{S}^*, h_\mathrm{T}^*) + S_\mathcal{H}^\ell(P_\mathrm{T}, P_\mathrm{S}).
\end{equation}
Another generalization error bound by \citet{mansour2009domain} can be obtained by upper bounding the S-disc to the worst hypothesis pair in the supremum term as follows:
\begin{equation}\label{bound_3}
R_{P_\mathrm{S}}^\ell(h,h_\mathrm{S}^*) + R_{P_\mathrm{T}}^\ell(h_\mathrm{S}^*, h_\mathrm{T}^*) + \mathrm{disc}_\mathcal{H}^\ell(P_\mathrm{T}, P_\mathrm{S}).
\end{equation}
\par
These bounds give \emph{sufficient} conditions for the success of domain adaptation with three terms: (\romannumeral 1) the expected loss with respect to $h_\mathrm{S}^*$ in the \emph{source domain}, (\romannumeral 2) the difference between $h_\mathrm{S}^*$ and $h_\mathrm{T}^*$ in the target domain and (\romannumeral 3) a discrepancy measure. 
Since we can minimize the first term, these bounds show that the domain adaptation will be successful if (\romannumeral 2) the infeasible term and (\romannumeral 3) the discrepancy measure are sufficiently small. 
In unsupervised domain adaptation, however, label information in the target domain is unknown which would be necessary to estimate (\romannumeral 2) the infeasible term.
\par
Here, a discrepancy measure is important since it assumes when the discrepancy measure is small, the difference between true risk minimizers in the two domains is also expected to be small \citep{mansour2009domain, kuroki2018unsupervised}. 
However, it is important to note that, strictly speaking, a small discrepancy measure does not \emph{always} guarantee whether adaptation can be successful without further assumptions~\citep{ben2010impossibility}.
A simple example is the covariate shift case where only two input distributions are slightly different but the true risk minimizers can be significantly different~\citep{sugiyama2007covariate}.
Nevertheless, there are many pieces of evidence that unsupervised domain adaptation works well in practice ~\citep{ganin2016domain, saito2017asymmetric, Saito_2018_CVPR}.
\par
Although introducing the supremum term in the  generalization error bound that considers the whole hypothesis class is interpretable and intuitive, it may induce a loose bound when a complex model is applied.
Moreover, due to the difficulty of learning with a complex model in practice, \emph{accurate} estimation of discrepancy measures based on the worst pair of hypotheses is difficult in both the binary and multiclass settings.

\section{Proposed Method}\label{sec:proposed}
In this section, we propose a novel discrepancy measure called \emph{paired hypotheses discrepancy} (PHD) to alleviate the above-mentioned limitations of the existing methods.
We also provide  generalization error bounds in the target domain based on PHD in Section \ref{sec:theory}. 

\subsection{Generalization error bound without the supremum term}
First, we define the general form of PHD as follows\footnote{We define it to avoid confusion between a discrepancy measure and other terms in the  generalization error bound.}:
\begin{definition}[Paired hypotheses discrepancy]\label{def:phd}
For any hypotheses $h_1 \in \mathcal{H}_1$ and $h_2 \in \mathcal{H}_2$ which are defined on the domain ($P_\mathrm{T}$, $f_\mathrm{T}$), paired hypotheses discrepancy (PHD) is defined as
\begin{equation*}
    \mathrm{PHD}_{P_\mathrm{T}}^\ell(h_1, h_2) = R_{P_\mathrm{T}}^\ell(h_1,h_2).
\end{equation*}
\end{definition}
Obviously, PHD can be computed in a straightforward manner without any approximations if a pair of hypotheses is given. Also it is symmetric $ R_{P_\mathrm{T}}^\ell(h_1, h_2) = R_{P_\mathrm{T}}^\ell(h_2, h_1)$ if the loss function $\ell$ is symmetric, $\ell(a,b)=\ell(b,a)$. Here, we propose another way to derive a  generalization error bound based on PHD and the triangle inequality as follows:
\begin{theorem}\label{bound_our}
Assume that $\ell$ obeys the triangle inequality, such as the zero-one loss. Then, for any hypothesis $h\in\mathcal{H}$,
\begin{equation}\label{eq1}
    R_{P_\mathrm{T}}^\ell(h, f_\mathrm{T}) - R_{P_\mathrm{T}}^\ell(h_\mathrm{T}^*, f_\mathrm{T}) \leq 
    R_{P_\mathrm{T}}^\ell(h,h_1) + R_{P_\mathrm{T}}^\ell(h_2, h_\mathrm{T}^*) + \mathrm{PHD}_{P_\mathrm{T}}^\ell(h_1, h_2).
\end{equation}
\begin{proof}\renewcommand{\qedsymbol}{}
Similarly to the inequality (\ref{bound}),
\begin{align*}
    R_{P_\mathrm{T}}^\ell(h, f_\mathrm{T}) &- R_{P_\mathrm{T}}^\ell(h_\mathrm{T}^*, f_\mathrm{T})  \\&\leq
    R_{P_\mathrm{T}}^\ell(h,h_1) + R_{P_\mathrm{T}}^\ell(h_1, h_\mathrm{T}^*)\numberthis{\label{ineq:stupid}}  \\ & \leq
    R_{P_\mathrm{T}}^\ell(h,h_1) + R_{P_\mathrm{T}}^\ell(h_1, h_2) + R_{P_\mathrm{T}}^\ell(h_2, h_\mathrm{T}^*). 
\end{align*}
\end{proof}
\end{theorem} 
\par
Unlike the existing work, where supremum terms have been introduced for their discrepancy measures \citep{ben2007analysis, mansour2009domain,kuroki2018unsupervised}, our bound in \eqref{eq1} is derived by using the triangle inequality again from  the bound in \eqref{bound}.
This bound also gives a \emph{sufficient} condition for the success of domain adaptation with three terms:  (\romannumeral 1) the expected loss with respect to $h_1$ in the \emph{target domain}, (\romannumeral 2) the difference between $h_2$ and $h_\mathrm{T}^*$ in the target domain which is an infeasible term and (\romannumeral 3) the discrepancy measure PHD. 
One may be tempted to choose $h_1=h_2$ since it makes PHD zero regardless of domains, which leads to Ineqs.~\eqref{bound} and \eqref{ineq:stupid}, but this cannot suggest any algorithm while our bound with different $h_1$ and $h_2$ can be informative as a discrepancy measure in unsupervised domain adaptation. 
In other words, we cannot make use of the bound when $h_1$ and $h_2$ are always same regardless of domains since the term $R_{P_\mathrm{T}}^\ell(h_2, h_\mathrm{T}^*)$ cannot be estimated, while the first term $R_{P_\mathrm{T}}^\ell(h,h_1)$ depends on a hypothesis $h$.

We can see that this bound holds for any hypotheses $h_1$ and $h_2$. 
Thus, the choice of $h_1$ and $h_2$ is critical for the tightness of the bound since it has to be \emph{informative} as a discrepancy measure between domains. 
We introduce one intuitive choice of the pair to make PHD informative in  Section \ref{ssl}.
Here, a discrepancy measure is said to be informative when it takes a small/large value for two similar/dissimilar domains.

Differently from the generalization error bound for the existing methods, this bound calculates the first term (\romannumeral 1), the expected loss with respect to $h_1$, in the \emph{target domain} which can be interpreted as pseudo-labeled target data based on hypothesis $h_1$.
Note that in the bound in Theorem \ref{bound_our}, we can deal with the infeasible term (\romannumeral 2) $R_{P_\mathrm{T}}^\ell(h_2, h_\mathrm{T}^*)$ in a more active way by using general hypothesis $h_2$ including $h_\mathrm{S}^*$. 
By incorporating unlabeled target data or prior knowledge into $h_2$, we can reduce $R_{P_\mathrm{T}}^\ell(h_2, h_\mathrm{T}^*)$ as shown empirically in Section \ref{sec:exp_infeasible}.

Even though the generalization error bound of PHD cannot be compared with that of the existing methods directly, we show that our bound is tighter than the existing methods empirically for neural networks in the experiment section.
This is intuitive since our bound does not contain the supremum term. 

\subsection{Choice of $h_1$ and $h_2$}\label{ssl}
Although the simple generalization error bound in Theorem \ref{bound_our} does not contain the supremum term as existing bounds, the choice of $h_1$ and $h_2$ is critical to make the bound informative and tight. This is when one can provide prior knowledge to the system.

Here, we suggest one intuitive example to pick a pair of $h_1$ and $h_2$ which does not need any prior knowledge on the task. First, we can choose one hypothesis based on semi-supervised learning where we are given a few labeled data and a large number of unlabeled data.
The main assumption behind this approach is that labeled data and unlabeled data are from the same domain and its objective is same as that of supervised methods: find a good hypothesis in the domain.
The main challenge in this approach is how to incorporate unlabeled data~\citep{chapelle2006SSL_MIT}.
It is difficult to use unlabeled data effectively without any additional assumptions~\citep{ben2008does,  singh2009unlabeled, darnstadt2013unlabeled}. 
Many assumptions have been considered in the literature of semi-supervised learning such as the continuity, manifold, and cluster assumptions~\citep{belkin2006manifold, chapelle2006SSL_MIT}.
Note that the performance of the model learned from the algorithm is known to be less desirable if the assumptions of the algorithm are violated~\citep{cozman2002unlabeled,  tian2004new, li2015towards, krijthe2017robust, sakai2017semi}.

Since we have a large number of labeled data in the source domain in unsupervised domain adaptation unlike standard semi-supervised learning, we suggest that if the source and target domains are similar, the hypothesis learned only from labeled data (supervised methods) should not be significantly different from the one learned from both labeled and unlabeled data (semi-supervised methods).
On the other hand, if the hypothesis becomes highly different when incorporating unlabeled data, this may indicate that the source and target domains are only weakly related or unrelated.
In this case, we may choose another hypothesis as the source hypothesis $h_\mathrm{S}^*$.


\begin{figure}[t]
\begin{center}
\centerline{\includegraphics[scale =0.7]{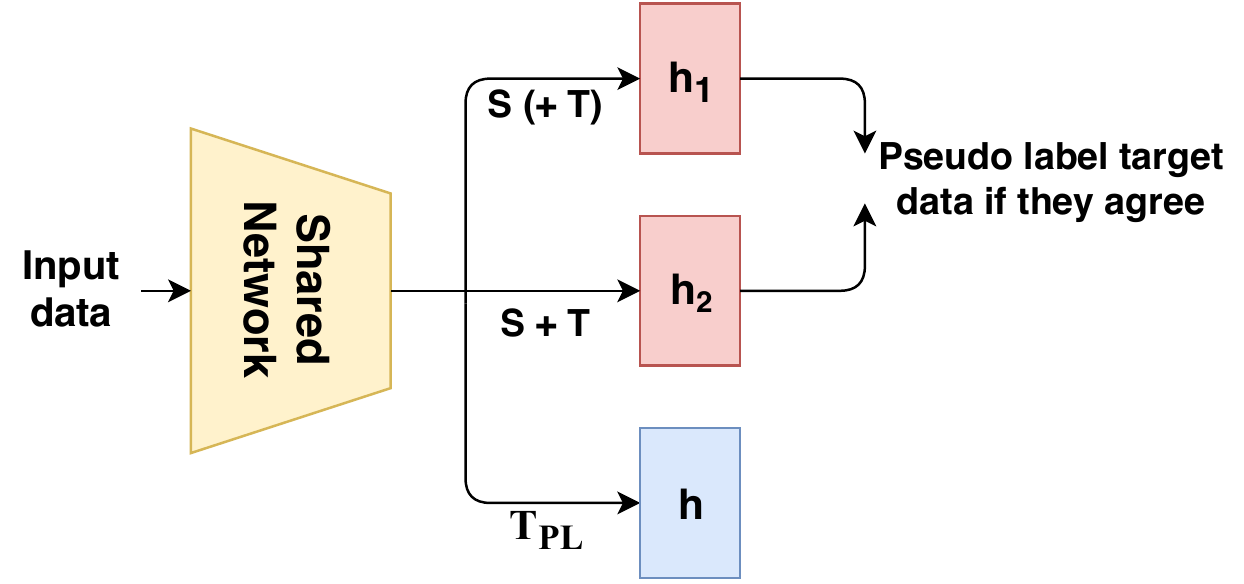}}
\caption{Asymmetric tri-training network proposed by \citet{saito2017asymmetric}. $\mathcal{S}$, $\mathcal{T}$ and $\mathcal{T}_{\mathrm{PL}}$ mean source data, target data and pseudo-labeled target data respectively.}
\vspace{-0.3in}
\label{fig:ete-network}
\end{center}
\end{figure}

\subsection{Application: end-to-end networks based on tri-training}\label{sec:saito}
Here, we show that the existing asymmetric tri-training network in \citet{saito2017asymmetric} can be interpreted as an end-to-end network that attempts to minimize PHD and find a hypothesis $h$ that minimizes our generalization error bound.
As illustrated in Figure \ref{fig:ete-network}, in the asymmetric tri-training setting, two models $h_1, h_2$ are trained independently using source data after passing the shared network and one model $h$ learns from pseudo-labeled target data by $h_1$ and $h_2$.
In the training stage, they construct the set of pseudo-labeled target data $\mathcal{T}_{\mathrm{PL}}\ \coloneqq \{(x_i, y_i) \vert \thickspace x_i \in \mathcal{T} \thickspace s.t. \thickspace h_1(x_i)=h_2(x_i) \coloneqq y_i\}$ only with the target data where two models $h_1$ and $h_2$ give same outputs.
Then, they learn the target domain with pseudo-labeled target data $\mathcal{T}_{\mathrm{PL}}$, which always satisfy $\widehat{\mathrm{PHD}}_{\mathrm{T}_{\mathrm{PL}}}^{\ell_{01}}(h_1,h_2)=0$.
After iterations, $\widehat{\mathrm{PHD}}_{\mathrm{T}_\mathrm{PL}}^\ell(h_1,h_2)$ converges to  $\mathrm{PHD}_{P_\mathrm{T}}^\ell(h_1,h_2)$ as the number of sampled target data increases.
Simultaneously, they train $h$ with pseudo-labeled target data, which indicates that they try to minimize the first term $R_{P_\mathrm{T}}(h,h_1)$ in the RHS of the generalization error bound in Theorem~\ref{bound_our}.
Note that the term $\hat{R}_{\mathrm{T}_\mathrm{PL}}(h,h_1)$ also converges to $R_{P_\mathrm{T}}(h,h_1)$ as the number of target data increases. 
More details are given in Theorem~\ref{thmsaito}.

\citet{saito2017asymmetric} explained that their tri-training network benefits from the discrepancy bound~\citep{mansour2009domain, ben2010theory}. 
However, we argue that this is not the case because the term they minimize is inside the supremum term of the discrepancy. 
Therefore, a pair $h_1$ and $h_2$ that minimizes the term in the supremum should be lower than the true supremum value.
Therefore, the discrepancy bound in~\citet{ben2010theory} is not useful for their asymmetric tri-training framework, while our generalization error bound based on PHD in Theorem~\ref{bound_our} can explain the success of the asymmetric tri-training network.

\section{Theoretical Analysis}\label{sec:theory}

In this section, we show an empirical estimator of PHD 
converges to the true PHD and can be accurately estimated from a finite number of data and provide an generalization error bound analysis.
To derive theoretical results of the generalization error bounds using PHD, the Rademacher complexity \citep{bartlett2002rademacher} is used, which measures the ability of a hypothesis class to fit random noise.
\begin{definition}[Rademacher complexity~\citep{bartlett2002rademacher}]\label{rademacher}
Let $\mathcal{H}$ be a set of real-valued hypotheses defined over a set $\mathcal{X}$. Given a sample ($x_1,\ldots,x_m$)$\in \mathcal{X}^m$ independently and identically drawn from a distribution $\mu$, the Rademacher complexity of $\mathcal{H}$ is defined as
\begin{equation*}
 \mathfrak{R}_{m}(\mathcal{H}) = \mathbb{E}_{x_1,\ldots,x_m}\mathbb{E}_{\sigma}\left[\sup\limits_{h\in\mathcal{H}}\left(\frac{1}{m}\sum_{i=1}^{m}\sigma_i h(x_i)\right)\right],
\end{equation*}
where the inner expectation is taken over $\sigma=(\sigma_1, \ldots , \sigma_m)$ which are independent random variables taking values in $\{-1, +1\}$. 
\end{definition}
The following theorem gives an upper bound of the deviation of the empirical PHD from the true PHD.
\begin{theorem}\label{thm:mainesb}
Let $\mathcal{H}$ be a set of hypotheses taking values in $\{-1, +1\}$. When we consider the zero-one loss $\ell$ in the binary classification, for empirical risk minimizers $\hat{h}_i$, any true risk minimizers $h_i^*$, for $i\in\{0,1\}$ respectively, in the hypothesis class $\mathcal{H}$, and any $\delta \in [0,1]$ the following inequality holds with probability at least $1-\delta$:
\begin{equation*}
    \mathrm{|}\widehat{\mathrm{PHD}}_\mathrm{T}^\ell(\hat{h}_1, \hat{h}_2) - \mathrm{PHD}_{P_\mathrm{T}}^\ell(h_1^*, h_2^*)\mathrm{|}  \leq 3 \mathfrak{R}_{n_\mathrm{T}}(\mathcal{H}) + 3\sqrt{\frac{\log(\frac{12}{\delta})}{2n_\mathrm{T}}} + \widehat{R}_{\mathrm{T}}^\ell(\hat{h}_1, h_1^*) +
    \widehat{R}_{\mathrm{T}}^\ell(\hat{h}_2, h_2^*).
\end{equation*}
\end{theorem}
The proof of this theorem is given in Appendix. 
Since this theorem holds for any true risk minimizers, the true risk minimizer does not need to be unique.
Rather, as an empirical risk minimizer approximates one of the true risk minimizers accurately through the learning procedure, the last two terms can be negligible.
From this theorem, we can see that the empirical PHD converges to the true PHD as the number of target data increases.

The following theorem shows the generalization bound of PHD in a finite sample case when we consider the zero-one loss in binary classification (its proof is given in Appendix). 

\begin{theorem}\label{thm:full}
Let $\mathcal{H}$ and $\mathcal{H}^\prime$ be a set of hypotheses taking values in $\{-1, +1\}$.
When we consider the zero-one loss $\ell$ in binary classification, then, for all $h \in \mathcal{H^\prime}$, empirical risk minimizers $\hat{h}_i$ and any true risk minimizers $h_i^*$ in the hypothesis class $\mathcal{H}$ for $i\in\{0,1\}$ respectively, all $\delta \in$ (0,1), then, with probability at least $1-\delta$,
\begin{multline*}
    R_{P_\mathrm{T}}^\ell(h,f_\mathrm{T})-R_{P_\mathrm{T}}^\ell(h_\mathrm{T}^*,f_\mathrm{T}) \\ \leq \widehat{R}_T^\ell(h, h_1^*) + \widehat{\mathrm{PHD}}_\mathrm{T}^\ell(\hat{h}_1, \hat{h}_2) + R_{P_\mathrm{T}}^\ell(h_2^*, h_\mathrm{T}^*) + 
    \mathfrak{R}_{n_\mathrm{T}}(\mathcal{H}^\prime) \\+ 3\mathfrak{R}_{n_\mathrm{T}}(\mathcal{H}) + 4\sqrt{\frac{\log\frac{7}{\delta}}{2n_\mathrm{T}}}
+\widehat{R}_{\mathrm{T}}^\ell(\hat{h}_1, h_1^*) +
    \widehat{R}_{\mathrm{T}}^\ell(\hat{h}_2, h_2^*).
\end{multline*}
\end{theorem}
This theorem considers the learning procedure of the pair of hypotheses in the estimation of PHD. 
However, the algorithm by \citet{saito2017asymmetric}, which was introduced in Section~\ref{sec:saito}, does not contain the estimation of PHD in the learning procedures of $h_1$ and $h_2$. 
This means that objective functions of hypotheses are not related to PHD. 
Instead, their algorithm samples the target data only when two fixed hypotheses agree, which makes PHD zero on the sampled target data even though it does not allow direct estimation of PHD.
The following theorem gives a generalization bound of PHD when we are given fixed hypotheses before the estimation, which can explain the behavior of the algorithms by \citet{saito2017asymmetric} (its proof is given in Appendix).

\begin{theorem}\label{thmsaito}
Let $\mathcal{H}$ be a set of hypotheses taking values in $\{-1, +1\}$.
Suppose that we consider the zero-one loss $\ell$ in binary classification and $h_1$ and $h_2$ are given. Then, for all $h \in \mathcal{H}$ and $\delta \in$ (0,1) with probability at least $1-\delta$,
\begin{equation*}
   \hspace{-0.15in} R_{P_\mathrm{T}}^\ell(h,f_\mathrm{T})-R_{P_\mathrm{T}}^\ell(h_\mathrm{T}^*,f_\mathrm{T}) \leq \widehat{R}_T^\ell(h, h_1) + \widehat{\mathrm{PHD}}_\mathrm{T}^\ell(h_1, h_2) + R_{P_\mathrm{T}}^\ell(h_2, h_\mathrm{T}^*) + 2\mathfrak{R}_{n_\mathrm{T}}(\mathcal{H}) + 2\sqrt{\frac{\log\frac{2}{\delta}}{2n_\mathrm{T}}}.
\end{equation*}
\end{theorem}
Extensions of the above results to multiclass problems are given in supplementary material.
Theorem~\ref{thmsaito} shows a generalization error bound in a finite sample case when two hypotheses are fixed a priori. In this case, the generalization error bound becomes tighter than Theorem~\ref{thm:full}, where the learning procedure of two hypotheses is contained.

\section{Experiments} \label{sec_exp}
In this section, we provide experiment results to demonstrate the usefulness of PHD. We pick $h_2:=h_\mathrm{SSL}$ based on semi-supervised learning as discussed in Section~\ref{ssl}.
We used two semi-supervised learning methods: semi-supervised binary classification based on positive-unlabeled classification (PNU)~\citep{sakai2017semi} for Section~\ref{subsec_cmpr}, and virtual adversarial training (VAT)~\citep{8417973} which can be used in both binary and multiclass classifications for
Sections~\ref{sec:exp_infeasible} and~\ref{sec:source_selection}. 
Note that target data used to train $h_\mathrm{SSL}$ were not used to calculate PHD to make sure that we calculated PHD with unseen data.
All experiment results are reported in the mean values with standard deviation over 10 trials. 
PHD was calculated with respect to the zero-one loss $\widehat{\mathrm{PHD}}_{\mathrm{T}}^{\ell_{01}}(h_\mathrm{S}^*, h_\mathrm{SSL})$ to be compared with existing methods.
Note that all values were calculated based on empirical risk minimizers $\widehat{h}$, but for simplicity, we denote it as $h$ in this section.
In all experiments, Adam~\citep{kingma2014adam} with AMSGRAD~\citep{reddi2018convergence} was used as an optimization algorithm. 
The learning rate was set at 0.001.
Note that $h_\mathrm{S}$ and $h_\mathrm{SSL}$ were obtained independently.
The details of implementation and datasets and more experimental results are given in Appendices B and C.
\begin{table*}[t]
\caption{Empirical generalization error bounds of PHD and $\mathrm{S}$-disc. Since the source and target domains are identical, discrepancy measures and infeasible terms are supposed to be small. The mean and standard deviation over 10 trials are reported. In the multiclass setting, existing methods \citep{ben2007analysis, mansour2009domain, kuroki2018unsupervised} cannot be computed to the best of our knowledge. We used the boldface for the best performance in a discrepancy measure and the infeasible term respectively.}
\label{tab:exp_bn}
\centering
\begin{small}
\begin{tabular}{ccccccc}
     \toprule
     Binary Setting&&&\multicolumn{2}{c}{Discrepancy Measure}&\multicolumn{2}{c}{Infeasible Term}\\
     \cmidrule{4-7}
       Dataset & Model & $h_\mathrm{S}$(\%) & $\mathrm{S}$-disc & $\mathrm{PHD}_\mathrm{VAT}$ &  $R_{\mathrm{T}}^{\ell_{01}}(h_\mathrm{VAT}, h_\mathrm{T}^*)$ & $R_{\mathrm{T}}^{\ell_{01}}(h_\mathrm{S}, h_\mathrm{T}^*)$  \\
    \midrule
     MNIST & MLP & 97.21 (0.124) & 0.6952 (0.186) & \textbf{0.0157 (0.025)} & \textbf{0.0176 (0.002)}  & 0.0210 (0.002) \\
    \midrule
      EMNIST & MLP & 99.03 (0.062) & 0.2379 (0.012) & \textbf{0.0069 (0.001)} & \textbf{0.0069 (0.000)} & 0.0086 (0.000) \\
     \midrule
     \multirow{2}{*}{CIFAR-10} & MLP & 72.53 (0.379) & 0.7220 (0.026) & \textbf{0.1496 (0.007)} & \textbf{0.2374 (0.011)} & 0.2461 (0.005) \\
      & VGG11 & 84.21 (0.543) & 0.9699 (0.004) & \textbf{0.1268 (0.009)} & \textbf{0.1385 (0.007)} & 0.1402 (0.005) \\ 
      \toprule
     Multiclass Setting&&&\multicolumn{2}{c}{Discrepancy Measure}&\multicolumn{2}{c}{Infeasible Term}\\
     \cmidrule{4-7}
       Dataset & Model & $h_\mathrm{S}$(\%) & \multicolumn{2}{c}{$\mathrm{PHD}_\mathrm{VAT}$} & $R_{\mathrm{T}}^{\ell_{01}}(h_\mathrm{VAT}, h_\mathrm{T}^*)$ & $R_{\mathrm{T}}^{\ell_{01}}(h_\mathrm{S}, h_\mathrm{T}^*)$  \\
    \midrule
      MNIST & MLP & 96.83 (0.313) & \multicolumn{2}{c}{0.0291 (0.004)} & \textbf{0.0305 (0.003)} & 0.0368 (0.003) \\
    \midrule
     CIFAR-10 & VGG11 & 73.66 (0.817) & \multicolumn{2}{c}{0.2223 (0.008)} & \textbf{0.2348 (0.133)} & 0.2415 (0.010) \\
      \midrule
      \multirow{2}{*}{SVHN} & MLP & 75.73 (0.601) & \multicolumn{2}{c}{0.1833 (0.006)} & \textbf{0.2246 (0.006)} & 0.2374 (0.006) \\
      & VGG11 & 90.39 (0.339) & \multicolumn{2}{c}{0.0852 (0.005)} & \textbf{0.0817 (0.006)} & 0.0898 (0.003) \\
      \bottomrule
\end{tabular}
\end{small}
\end{table*}
\subsection{Comparison with existing methods} \label{subsec_cmpr}
We first explain the notations used in the results reported in Tables~\ref{tab:exp_cmp} and~\ref{tab:exp_bn}.
$h_\mathrm{S}$(\%) denotes the accuracy of empirical risk minimizer $h_\mathrm{S}$ with respect to the true labeling function $f_\mathrm{S}$ in the source domain. 
Clearly, the performance of $h_\mathrm{S}$ is crucial for both PHD and $\mathrm{S}$-disc.
Here, to compare with existing methods which are limited to the binary setting, we used the MNIST \citep{lecun1998mnist} and EMNIST (extended MNIST) \citep{cohen2017emnist} datasets divided into the odd and even numbers.
Note that EMNIST is the dataset that has 4 times more data than the original MNIST dataset.
\\
\noindent \textbf{Linear Model:} For linear models, existing discrepancy measures and PHD worked properly in that the values of discrepancy measure terms were small when the source and target domains are identical as illustrated Table \ref{tab:exp_cmp}. This result shows these discrepancy measures are useful for a simple hypothesis class.
\\
\noindent \textbf{MLPs Model:} It is observed from Table \ref{tab:exp_cmp} that when the complex models are applied, estimated values of existing methods were large although the source and target domains are identical. This result shows that even when the source and target domains are identical, in the complex model, we may try to adapt the \emph{same} domains unnecessarily if we used existing methods.

\subsection{Observation of the term $R_{P_\mathrm{T}}^\ell(h_2, h_\mathrm{T}^*)$ and the tightness of the empirical generalization error bound} \label{sec:exp_infeasible}
Here, we take a closer look at the infeasible term $R_{P_\mathrm{T}}^\ell(h_2, h_\mathrm{T}^*)$ in the generalization error bound in the target domain.
In unsupervised domain adaptation, we cannot estimate this term since there is no information to learn the target hypothesis $h_\mathrm{T}^*$. 
Therefore, we calculate the infeasible term using target labels to \emph{illustrate} the value of the infeasible term for all methods. 
Note that the target labels are not given to any methods.
By having both the infeasible term and discrepancy measure term, the generalization error bound in the target domain can be calculated \emph{empirically}. 
Moreover, the first term in the RHS of the generalization error bound in \eqref{bound_2} and \eqref{eq1} can be minimized by training  $h$ and thus we ignored this term when comparing the bound.
To calculate $h_\mathrm{T}^*$, we divided train data of each dataset into disjoint subsets.
Note that true values of each discrepancy measure and infeasible terms are zero when two domains identical, however, because of its difficulty in estimation, existing method gives unreliable values in the complex model.

\noindent \textbf{Binary:} As illustrated in Tables \ref{tab:exp_cmp} and \ref{tab:exp_bn}, PHD was always smaller than $\mathrm{S}$-disc when they were applied to complex models. In most cases, $R_{\mathrm{T}}^{\ell_{01}}(h_\mathrm{VAT}, h_\mathrm{T}^*)$ was smaller than $R_{\mathrm{T}}^{\ell_{01}}(h_\mathrm{S}, h_\mathrm{T}^*)$. 
By the generalization error bound in Ineq. (\ref{bound_2}) and Theorem \ref{bound_our}, PHD always provided the tighter bound empirically than $\mathrm{S}$-disc and the discrepancy distance when the complex models such as MLPs and CNNs were applied.
\\
\noindent \textbf{Multiclass:} Unlike existing methods, PHD can be easily estimated in the multiclass setting since it only considers two hypotheses and has no supremum term.
From Table \ref{tab:exp_bn}, PHD was observed to be stable for binary and multiclass settings since it outputted similar values for both settings. 
In most experiments for the multiclass setting, $R_T^{\ell_{01}}(h_\mathrm{VAT}, h_\mathrm{T}^*)$ shows a slightly better performance than $R_T^{\ell_{01}}(h_\mathrm{S}, h_\mathrm{T}^*)$.
\\
\noindent \textbf{Different domains:} As a discrepancy measure, PHD should become small when two domains are similar and large when two domains are different. From the Table~\ref{tab:exp_diff}, we can see that the value of PHD becomes large when two domains are different. More specifically, its values are close to 0.9 which can be obtained by random guess when the number of classes is 10. It could be interpreted as the pair of hypotheses are unrelated.


\begin{table*}[t]
\caption{Empirical generalization error bound of PHD in the multiclass setting with VGG11 model}
\label{tab:exp_diff}
\begin{center}
\begin{small}
\begin{tabular}{cccc|cc}
     \toprule
     &&&&\multicolumn{2}{c}{Infeasible Term} \\
       Source & Target & $h_\mathrm{S}$(\%) & $\mathrm{PHD}_\mathrm{VAT}$ & $R_{P_\mathrm{T}}^{\ell_{01}}(h_\mathrm{VAT}, h_\mathrm{T}^*)$ & $R_{P_\mathrm{T}}^{\ell_{01}}(h_\mathrm{S}, h_\mathrm{T}^*)$  \\
    \midrule
       \multirow{3}{*}{MNISTM} & MNISTM& 95.10 (0.323) & 0.0501 (0.005) & \textbf{0.0436 (0.004)} & 0.0540 (0.005) \\
      &SVHN & 95.18 (0.384) & 0.8589 (0.072) & 0.8693 (0.045) & 0.5968 (0.008) \\
      &CIFAR10& 95.03 (0.349)  & 0.8960 (0.022) & 0.9007 (0.005)  & 0.9223 (0.006) \\ 
      \bottomrule
\end{tabular}
\end{small}
\end{center}
\vspace{-0.2in}
\end{table*}
\begin{figure}[h]
\begin{center}
\centerline{\includegraphics[scale=0.65]{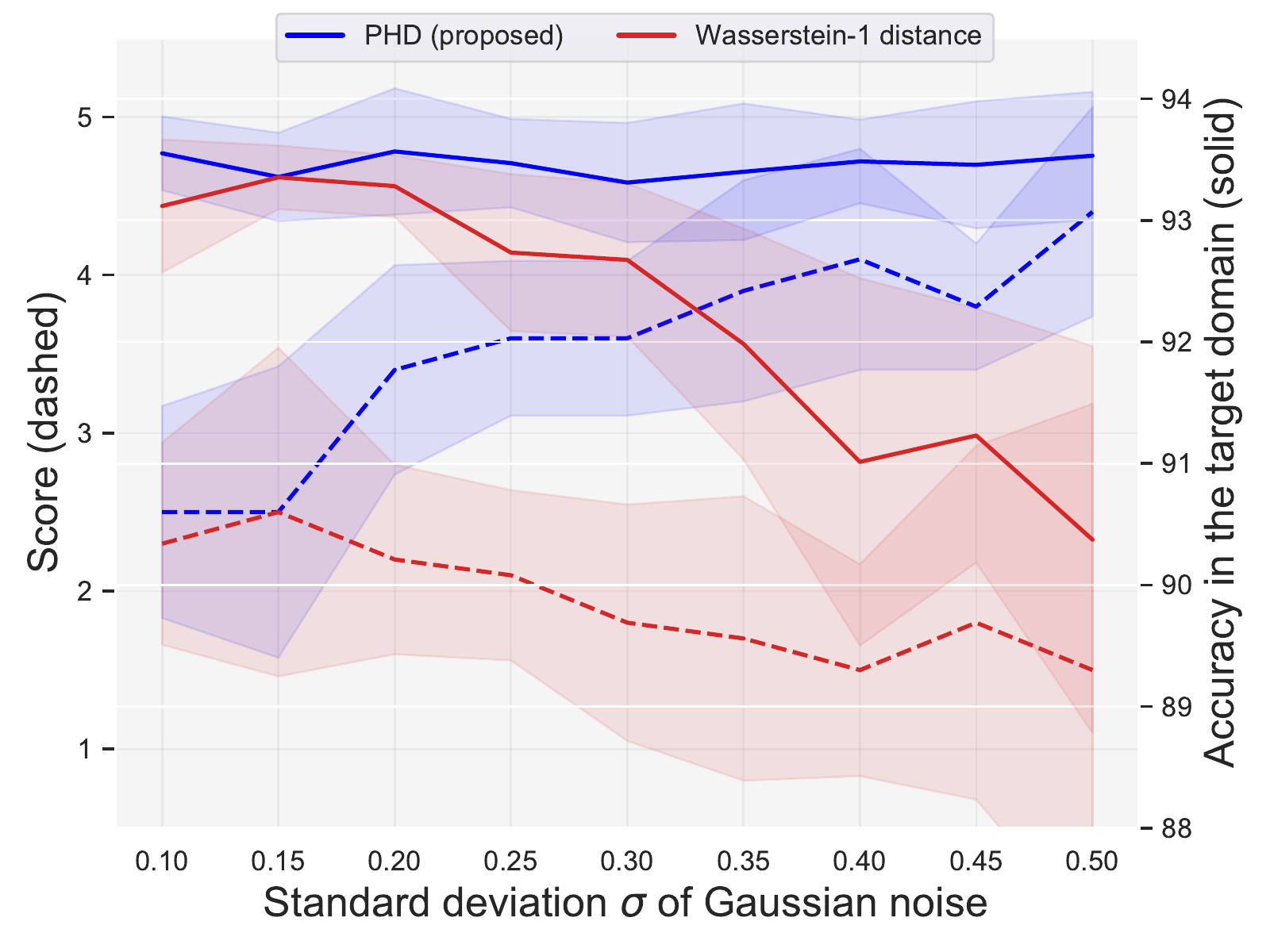}}
\vspace{-0.15in}
\caption{Source selection score (dashed line) and accuracy (solid line) in the target domain with selected source after adapted by correlation alignment (CORAL)~\citep{sun2016return} in the multiclass setting. Higher values are better.}
\vspace{-0.5in}
\label{fig:source_selection}
\end{center}
\end{figure}
\par
\subsection{Source selection} \label{sec:source_selection}
Finally, we show the performance of PHD and the Wasserstein-1 (W1) distance~\citep{redko2017theoretical} in the source selection task in the \emph{multiclass setting} where existing methods~\citep{ben2007analysis,mansour2009domain,kuroki2018unsupervised} cannot be used. 
The experiment was designed similarly to the one that was considered by~\citet{kuroki2018unsupervised}, which is an experiment to select a better source.
The goal of source selection is to correctly rank clean source domains over noisy source domains based on discrepancy measures in multiclass setting. 
The score indicates how many clean sources are ranked within the top-5 rank out of 10 sources (5 sources are clean among them). 
The accuracy indicates how well a classifier trained on data in the selected sources with the simple adaptation method, correlation alignment (CORAL)~\citep{sun2016return}, classifies the target data.

In this experiment, we used grayscale MNIST-M as the five clean source domains, grayscale MNIST-M corrupted by Gaussian random noise as five noisy source domains and MNIST as the target domain. We used the Gaussian noise with different standard deviations $\sigma$ for the noisy source domains. 
The value of mean and standard deviation of grayscale MNIST-M data are (0.4591, 0.2352) which means that noise over 0.55 might be meaningless since the value of each pixel could easily exceed 1.0 which is unrealistic to the grayscale image. 
5-layer MLP model with batch normalization \citep{ioffe2015batch} are used to calculate PHD while we calculated the W1 distance \citep{redko2017theoretical} by a Lipschitz function using the gradient penalty with the 5-layer MLP model without batch normalization.

Figure \ref{fig:source_selection} shows the score and accuracy of each discrepancy measure with different noise rates. 
PHD achieved a better performance as the noise rate increases. 
However, the W1 distance could not distinguish between noisy and clean source domains effectively when the small noise is applied. 
Moreover, the performance of W1 distance was not improved significantly as noise increased, which incurred accuracy drops in the target domain.
Note that the W1 distance does not take the hypothesis class into account and also cannot utilize the source labels, which can be less desirable.
On the other hand, PHD exploits the label information in the source domains and considers a pair of fixed hypotheses.
This might be the reason why PHD outperformed the W1 distance in this experiment.

\section{Conclusion}
We proposed a novel discrepancy measure for unsupervised domain adaptation called paired hypotheses discrepancy (PHD), which can be applied effectively for complex models such as deep neural networks.
The key idea is that PHD only considers a pair of hypotheses, not the whole hypothesis class. 
Furthermore, we derived generalization error bounds for our proposed method. 
We also showed that PHD can influence some algorithms in domain adaptation.
Finally, we show that PHD can be estimated effectively in both the binary and multiclass settings and demonstrated its usefulness in our experiments.


\section*{Acknowledgement}
We thank Han Bao, Hideaki Imamura, Masahiro Kato and anonymous Reviewers for helpful discussion. NC was supported by MEXT scholarship. MS was supported by KAKENHI 17H01760.

\bibliography{refs}

\begin{thebibliography}{55}
\providecommand{\natexlab}[1]{#1}
\providecommand{\url}[1]{\texttt{#1}}
\expandafter\ifx\csname urlstyle\endcsname\relax
  \providecommand{\doi}[1]{doi: #1}\else
  \providecommand{\doi}{doi: \begingroup \urlstyle{rm}\Url}\fi

\bibitem[Arjovsky et~al.(2017)Arjovsky, Chintala, and
  Bottou]{arjovsky2017wasserstein}
Martin Arjovsky, Soumith Chintala, and L{\'e}on Bottou.
\newblock Wasserstein {GAN}.
\newblock \emph{arXiv preprint arXiv:1701.07875}, 2017.

\bibitem[Bartlett and Mendelson(2002)]{bartlett2002rademacher}
Peter~L Bartlett and Shahar Mendelson.
\newblock Rademacher and {Gaussian} complexities: Risk bounds and structural
  results.
\newblock \emph{JMLR}, 2002.

\bibitem[Belkin et~al.(2006)Belkin, Niyogi, and Sindhwani]{belkin2006manifold}
Mikhail Belkin, Partha Niyogi, and Vikas Sindhwani.
\newblock Manifold regularization: A geometric framework for learning from
  labeled and unlabeled examples.
\newblock \emph{JMLR}, 2006.

\bibitem[Ben-David et~al.(2007)Ben-David, Blitzer, Crammer, and
  Pereira]{ben2007analysis}
Shai Ben-David, John Blitzer, Koby Crammer, and Fernando Pereira.
\newblock Analysis of representations for domain adaptation.
\newblock In \emph{NeurIPS}, 2007.

\bibitem[Ben-David et~al.(2008)Ben-David, Lu, and P{\'a}l]{ben2008does}
Shai Ben-David, Tyler Lu, and D{\'a}vid P{\'a}l.
\newblock Does unlabeled data provably help? worst-case analysis of the sample
  complexity of semi-supervised learning.
\newblock In \emph{COLT}, 2008.

\bibitem[Ben-David et~al.(2010{\natexlab{a}})Ben-David, Blitzer, Crammer,
  Kulesza, Pereira, and Vaughan]{ben2010theory}
Shai Ben-David, John Blitzer, Koby Crammer, Alex Kulesza, Fernando Pereira, and
  Jennifer~Wortman Vaughan.
\newblock A theory of learning from different domains.
\newblock \emph{Machine Learning}, 2010{\natexlab{a}}.

\bibitem[Ben-David et~al.(2010{\natexlab{b}})Ben-David, Lu, Luu, and
  P{\'a}l]{ben2010impossibility}
Shai Ben-David, Tyler Lu, Teresa Luu, and D{\'a}vid P{\'a}l.
\newblock Impossibility theorems for domain adaptation.
\newblock In \emph{AISTATS}, 2010{\natexlab{b}}.

\bibitem[Bousmalis et~al.(2017)Bousmalis, Silberman, Dohan, Erhan, and
  Krishnan]{bousmalis2017unsupervised}
Konstantinos Bousmalis, Nathan Silberman, David Dohan, Dumitru Erhan, and Dilip
  Krishnan.
\newblock Unsupervised pixel-level domain adaptation with generative
  adversarial networks.
\newblock In \emph{CVPR}, 2017.

\bibitem[Chapelle et~al.(2006)Chapelle, Schölkopf, and
  Zien]{chapelle2006SSL_MIT}
Olivier Chapelle, Bernhard Schölkopf, and Alexander Zien.
\newblock \emph{Semi-Supervised Learning}.
\newblock 2006.

\bibitem[Cohen et~al.(2017)Cohen, Afshar, Tapson, and van
  Schaik]{cohen2017emnist}
Gregory Cohen, Saeed Afshar, Jonathan Tapson, and Andr{\'e} van Schaik.
\newblock {EMNIST}: an extension of mnist to handwritten letters.
\newblock \emph{arXiv preprint arXiv:1702.05373}, 2017.

\bibitem[Cortes et~al.(2019)Cortes, Mohri, and Medina]{cortes2019adaptation}
Corinna Cortes, Mehryar Mohri, and Andr{\'e}s~Munoz Medina.
\newblock Adaptation based on generalized discrepancy.
\newblock \emph{JMLR}, 2019.

\bibitem[Courty et~al.(2017)Courty, Flamary, Habrard, and
  Rakotomamonjy]{courty2017joint}
Nicolas Courty, R{\'e}mi Flamary, Amaury Habrard, and Alain Rakotomamonjy.
\newblock Joint distribution optimal transportation for domain adaptation.
\newblock In \emph{NeurIPS}, 2017.

\bibitem[Cozman et~al.(2002)Cozman, Cohen, and Cirelo]{cozman2002unlabeled}
Fabio~Gagliardi Cozman, Ira Cohen, and M~Cirelo.
\newblock Unlabeled data can degrade classification performance of generative
  classifiers.
\newblock In \emph{Flairs Conference}, 2002.

\bibitem[Darnst{\"a}dt et~al.(2013)Darnst{\"a}dt, Simon, and
  Sz{\"o}r{\'e}nyi]{darnstadt2013unlabeled}
Malte Darnst{\"a}dt, Hans~Ulrich Simon, and Bal{\'a}zs Sz{\"o}r{\'e}nyi.
\newblock Unlabeled data does provably help.
\newblock In \emph{International Symposium on Theoretical Aspects of Computer
  Science}, 2013.

\bibitem[Deng et~al.(2014)Deng, Zhang, Eyben, and
  Schuller]{deng2014autoencoder}
Jun Deng, Zixing Zhang, Florian Eyben, and Bj{\"o}rn Schuller.
\newblock Autoencoder-based unsupervised domain adaptation for speech emotion
  recognition.
\newblock \emph{IEEE Signal Processing Letters}, 2014.

\bibitem[Ganin and Lempitsky(2015)]{pmlr-v37-ganin15}
Yaroslav Ganin and Victor Lempitsky.
\newblock Unsupervised domain adaptation by backpropagation.
\newblock In \emph{ICML}, 2015.

\bibitem[Ganin et~al.(2016)Ganin, Ustinova, Ajakan, Germain, Larochelle,
  Laviolette, Marchand, and Lempitsky]{ganin2016domain}
Yaroslav Ganin, Evgeniya Ustinova, Hana Ajakan, Pascal Germain, Hugo
  Larochelle, Fran{\c{c}}ois Laviolette, Mario Marchand, and Victor Lempitsky.
\newblock Domain-adversarial training of neural networks.
\newblock \emph{JMLR}, 2016.

\bibitem[Germain et~al.(2013)Germain, Habrard, Laviolette, and
  Morvant]{pmlr-v28-germain13}
Pascal Germain, Amaury Habrard, François Laviolette, and Emilie Morvant.
\newblock A pac-bayesian approach for domain adaptation with specialization to
  linear classifiers.
\newblock In \emph{ICML}, 2013.

\bibitem[Hinton et~al.(2012)Hinton, Deng, Yu, Dahl, Mohamed, Jaitly, Senior,
  Vanhoucke, Nguyen, Sainath, et~al.]{hinton2012deep}
Geoffrey Hinton, Li~Deng, Dong Yu, George~E Dahl, Abdel-Rahman Mohamed, Navdeep
  Jaitly, Andrew Senior, Vincent Vanhoucke, Patrick Nguyen, Tara~N Sainath,
  et~al.
\newblock Deep neural networks for acoustic modeling in speech recognition: The
  shared views of four research groups.
\newblock \emph{IEEE Signal Processing Magazine}, 2012.

\bibitem[Huang et~al.(2007)Huang, Gretton, Borgwardt, Sch{\"o}lkopf, and
  Smola]{huang2007correcting}
Jiayuan Huang, Arthur Gretton, Karsten~M Borgwardt, Bernhard Sch{\"o}lkopf, and
  Alex~J Smola.
\newblock Correcting sample selection bias by unlabeled data.
\newblock In \emph{NeurIPS}, 2007.

\bibitem[Ioffe and Szegedy(2015)]{ioffe2015batch}
Sergey Ioffe and Christian Szegedy.
\newblock Batch normalization: Accelerating deep network training by reducing
  internal covariate shift.
\newblock In \emph{ICML}, 2015.

\bibitem[Jiang and Zhai(2007)]{jiang2007instance}
Jing Jiang and ChengXiang Zhai.
\newblock Instance weighting for domain adaptation in {NLP}.
\newblock In \emph{ACL}, 2007.

\bibitem[Kingma and Ba(2014)]{kingma2014adam}
Diederik~P Kingma and Jimmy Ba.
\newblock Adam: A method for stochastic optimization.
\newblock \emph{arXiv preprint arXiv:1412.6980}, 2014.

\bibitem[Kodirov et~al.(2015)Kodirov, Xiang, Fu, and
  Gong]{kodirov2015unsupervised}
Elyor Kodirov, Tao Xiang, Zhenyong Fu, and Shaogang Gong.
\newblock Unsupervised domain adaptation for zero-shot learning.
\newblock In \emph{ICCV}, 2015.

\bibitem[Krijthe and Loog(2017)]{krijthe2017robust}
Jesse~H Krijthe and Marco Loog.
\newblock Robust semi-supervised least squares classification by implicit
  constraints.
\newblock \emph{Pattern Recognition}, 2017.

\bibitem[Krizhevsky et~al.(2012)Krizhevsky, Sutskever, and
  Hinton]{krizhevsky2012imagenet}
Alex Krizhevsky, Ilya Sutskever, and Geoffrey~E Hinton.
\newblock Image{Net} classification with deep convolutional neural networks.
\newblock In \emph{NeurIPS}, 2012.

\bibitem[Krizhevsky et~al.(2014)Krizhevsky, Nair, and
  Hinton]{krizhevsky2014cifar}
Alex Krizhevsky, Vinod Nair, and Geoffrey Hinton.
\newblock The {CIFAR-10} dataset.
\newblock \url{http://www.cs.toronto.edu/kriz/cifar.html}, 2014.

\bibitem[Kuroki et~al.(2019)Kuroki, Charoenphakdee, Bao, Honda, Sato, and
  Sugiyama]{kuroki2018unsupervised}
Seiichi Kuroki, Nontawat Charoenphakdee, Han Bao, Junya Honda, Issei Sato, and
  Masashi Sugiyama.
\newblock Unsupervised domain adaptation based on source-guided discrepancy.
\newblock In \emph{AAAI}, 2019.

\bibitem[Kuznetsov et~al.(2015)Kuznetsov, Mohri, and
  Syed]{kuznetsov2015rademacher}
Vitaly Kuznetsov, Mehryar Mohri, and U~Syed.
\newblock Rademacher complexity margin bounds for learning with a large number
  of classes.
\newblock In \emph{ICML Workshop on Extreme Classification: Learning with a
  Very Large Number of Labels}, 2015.

\bibitem[LeCun(1998)]{lecun1998mnist}
Yann LeCun.
\newblock The {MNIST} database of handwritten digits.
\newblock \url{http://yann.lecun.com/exdb/mnist/}, 1998.

\bibitem[Li and Zhou(2015)]{li2015towards}
Yu-Feng Li and Zhi-Hua Zhou.
\newblock Towards making unlabeled data never hurt.
\newblock \emph{IEEE Transactions on Pattern Analysis and Machine
  Intelligence}, 2015.

\bibitem[Long et~al.(2016)Long, Zhu, Wang, and Jordan]{long2016unsupervised}
Mingsheng Long, Han Zhu, Jianmin Wang, and Michael~I Jordan.
\newblock Unsupervised domain adaptation with residual transfer networks.
\newblock In \emph{NeurIPS}, 2016.

\bibitem[Maas et~al.(2013)Maas, Hannun, and Ng]{maas2013rectifier}
Andrew~L Maas, Awni~Y Hannun, and Andrew~Y Ng.
\newblock Rectifier nonlinearities improve neural network acoustic models.
\newblock In \emph{ICML}, 2013.

\bibitem[Mansour et~al.(2009{\natexlab{a}})Mansour, Mohri, and
  Rostamizadeh]{mansour2009domain}
Yishay Mansour, Mehryar Mohri, and Afshin Rostamizadeh.
\newblock Domain adaptation: Learning bounds and algorithms.
\newblock In \emph{COLT}, 2009{\natexlab{a}}.

\bibitem[Mansour et~al.(2009{\natexlab{b}})Mansour, Mohri, and
  Rostamizadeh]{mansour2009multiple}
Yishay Mansour, Mehryar Mohri, and Afshin Rostamizadeh.
\newblock Multiple source adaptation and the r{\'e}nyi divergence.
\newblock In \emph{UAI}, 2009{\natexlab{b}}.

\bibitem[McDiarmid(1989)]{mcdiarmid1989method}
Colin McDiarmid.
\newblock On the method of bounded differences.
\newblock \emph{Surveys in combinatorics}, 1989.

\bibitem[Miyato et~al.(2018)Miyato, Maeda, Ishii, and Koyama]{8417973}
T.~Miyato, S.~Maeda, S.~Ishii, and M.~Koyama.
\newblock Virtual adversarial training: A regularization method for supervised
  and semi-supervised learning.
\newblock \emph{IEEE Transactions on Pattern Analysis and Machine
  Intelligence}, 2018.

\bibitem[Mohri et~al.(2012)Mohri, Rostamizadeh, and
  Talwalkar]{Mohri:2012:FML:2371238}
Mehryar Mohri, Afshin Rostamizadeh, and Ameet Talwalkar.
\newblock \emph{Foundations of Machine Learning}.
\newblock 2012.

\bibitem[Nair and Hinton(2010)]{nair2010rectified}
Vinod Nair and Geoffrey~E Hinton.
\newblock Rectified linear units improve restricted boltzmann machines.
\newblock In \emph{ICML}, 2010.

\bibitem[Netzer et~al.(2011)Netzer, Wang, Coates, Bissacco, Wu, and
  Ng]{netzer2011reading}
Yuval Netzer, Tao Wang, Adam Coates, Alessandro Bissacco, Bo~Wu, and Andrew~Y
  Ng.
\newblock Reading digits in natural images with unsupervised feature learning.
\newblock In \emph{NeurIPS workshop on deep learning and unsupervised feature
  learning}, 2011.

\bibitem[Pan et~al.(2010)Pan, Yang, et~al.]{pan2010survey}
Sinno~Jialin Pan, Qiang Yang, et~al.
\newblock A survey on transfer learning.
\newblock \emph{IEEE Transactions on knowledge and data engineering}, 2010.

\bibitem[Radford et~al.(2015)Radford, Metz, and
  Chintala]{radford2015unsupervised}
Alec Radford, Luke Metz, and Soumith Chintala.
\newblock Unsupervised representation learning with deep convolutional
  generative adversarial networks.
\newblock \emph{arXiv preprint arXiv:1511.06434}, 2015.

\bibitem[Reddi et~al.(2018)Reddi, Kale, and Kumar]{reddi2018convergence}
Sashank~J Reddi, Satyen Kale, and Sanjiv Kumar.
\newblock On the convergence of adam and beyond.
\newblock 2018.

\bibitem[Redko et~al.(2017)Redko, Habrard, and Sebban]{redko2017theoretical}
Ievgen Redko, Amaury Habrard, and Marc Sebban.
\newblock Theoretical analysis of domain adaptation with optimal transport.
\newblock In \emph{ECML PKDD}, 2017.

\bibitem[Saito et~al.(2017)Saito, Ushiku, and Harada]{saito2017asymmetric}
Kuniaki Saito, Yoshitaka Ushiku, and Tatsuya Harada.
\newblock Asymmetric tri-training for unsupervised domain adaptation.
\newblock In \emph{ICML}, 2017.

\bibitem[Saito et~al.(2018)Saito, Watanabe, Ushiku, and
  Harada]{Saito_2018_CVPR}
Kuniaki Saito, Kohei Watanabe, Yoshitaka Ushiku, and Tatsuya Harada.
\newblock Maximum classifier discrepancy for unsupervised domain adaptation.
\newblock In \emph{CVPR}, 2018.

\bibitem[Sakai et~al.(2017)Sakai, Plessis, Niu, and Sugiyama]{sakai2017semi}
Tomoya Sakai, Marthinus~Christoffel Plessis, Gang Niu, and Masashi Sugiyama.
\newblock Semi-supervised classification based on classification from positive
  and unlabeled data.
\newblock In \emph{ICML}, 2017.

\bibitem[Shen et~al.(2018)Shen, Qu, Zhang, and Yu]{AAAI1817155}
Jian Shen, Yanru Qu, Weinan Zhang, and Yong Yu.
\newblock Wasserstein distance guided representation learning for domain
  adaptation.
\newblock In \emph{AAAI}, 2018.

\bibitem[Silver et~al.(2016)Silver, Huang, Maddison, Guez, Sifre, Van
  Den~Driessche, Schrittwieser, Antonoglou, Panneershelvam, Lanctot,
  et~al.]{silver2016mastering}
David Silver, Aja Huang, Chris~J Maddison, Arthur Guez, Laurent Sifre, George
  Van Den~Driessche, Julian Schrittwieser, Ioannis Antonoglou, Veda
  Panneershelvam, Marc Lanctot, et~al.
\newblock Mastering the game of go with deep neural networks and tree search.
\newblock \emph{nature}, 2016.

\bibitem[Singh et~al.(2009)Singh, Nowak, and Zhu]{singh2009unlabeled}
Aarti Singh, Robert Nowak, and Jerry Zhu.
\newblock Unlabeled data: Now it helps, now it doesn't.
\newblock In \emph{NeurIPS}, 2009.

\bibitem[Sugiyama et~al.(2007)Sugiyama, Krauledat, and
  M\"uller]{sugiyama2007covariate}
Masashi Sugiyama, Matthias Krauledat, and Klaus-Robert M\"uller.
\newblock Covariate shift adaptation by importance weighted cross validation.
\newblock \emph{JMLR}, 2007.

\bibitem[Sugiyama et~al.(2008)Sugiyama, Nakajima, Kashima, Buenau, and
  Kawanabe]{sugiyama2008direct}
Masashi Sugiyama, Shinichi Nakajima, Hisashi Kashima, Paul~V Buenau, and
  Motoaki Kawanabe.
\newblock Direct importance estimation with model selection and its application
  to covariate shift adaptation.
\newblock In \emph{NeurIPS}, 2008.

\bibitem[Sun et~al.(2016)Sun, Feng, and Saenko]{sun2016return}
Baochen Sun, Jiashi Feng, and Kate Saenko.
\newblock Return of frustratingly easy domain adaptation.
\newblock In \emph{AAAI}, 2016.

\bibitem[Tian et~al.(2004)Tian, Yu, Xue, and Sebe]{tian2004new}
Qi~Tian, Jie Yu, Qing Xue, and Nicu Sebe.
\newblock A new analysis of the value of unlabeled data in semi-supervised
  learning for image retrieval.
\newblock In \emph{ICME}, 2004.

\bibitem[Zhang et~al.(2012)Zhang, Zhang, and Ye]{zhang2012generalization}
Chao Zhang, Lei Zhang, and Jieping Ye.
\newblock Generalization bounds for domain adaptation.
\newblock In \emph{NeurIPS}, 2012.

\end{thebibliography}
\bibliographystyle{plainnat}

\newpage
\appendix
\onecolumn
\section{Proofs}
\subsection{Theorems used in proofs}
\begin{theorem}[Theorem 3.2 in \citet{Mohri:2012:FML:2371238}]\label{thm:bnry_rad}
Let $\mathcal{H}$ be a family of functions taking values in $\{-1, +1\}$ over a sample $\mathrm{T}$ of size $n_\mathrm{T}$ drawn according to $P_\mathrm{T}$, $h_1$ is given and $\ell$ is the zero-one loss function. Then, for any $\delta\in(0,1)$, with probability at least $1-\delta$, the following inequality holds for all $h\in\mathcal{H}$:
\begin{equation*}
    \mathrm{|}R_{P_\mathrm{T}}^\ell(h,h_1) - \widehat{R}^\ell_\mathrm{T}(h,h_1)\mathrm{|} \leq \mathfrak{R}_{n_\mathrm{T}}(\mathcal{H}) + \sqrt{\frac{\log\frac{2}{\delta}}{2n_\mathrm{T}}}.
\end{equation*}
\end{theorem}

\begin{theorem}[Theorem 2 in \citet{kuznetsov2015rademacher}]\label{thm:mlt_rad}
Let $\mathcal{H}$ be a family of hypotheses mapping $\mathcal{X}\times\mathcal{Y}$ to $\mathbb{R}$, with $\mathcal{Y}=\{1,\ldots,k\}$ in k-class classification. Fix $\rho\ge0$ and $h_1$ is given. Then, for any $\delta\in(0,1)$, with probability at least $1-\delta$, the following inequality holds for all $h\in\mathcal{H}$:
\begin{equation*}
     R_{P_\mathrm{T}}^{\ell}(h,f_\mathrm{T}) \leq \widehat{R}^{\ell_\rho}_{\mathrm{T}}(h, h_1) + \frac{4k}{\rho}\mathfrak{R}_{n_\mathrm{T}}(\Pi_1(\mathcal{H})) + \sqrt{\frac{\log{\frac{1}{\delta}}}{2n_\mathrm{T}}},
\end{equation*}
where $\Pi_1(\mathcal{H}) = \{(x, y) \rightarrow h(x,y) : y \in \mathcal{Y}$, $h \in \mathcal{H}\}$.
\end{theorem}


\subsection{Proof of Theorem \ref{thm:mainesb}}
\begin{proof}
By applying the triangle inequality, we could get two inequalities:
\begin{align*}
    \mathrm{|}\widehat{\mathrm{PHD}}_\mathrm{T}^\ell(\hat{h}_1, \hat{h}_2) - &\mathrm{PHD}_{P_\mathrm{T}}^\ell(h_1^*, h_2^*)\mathrm{|} \\&\leq
    \mathrm{|}\widehat{\mathrm{PHD}}_\mathrm{T}^\ell(\hat{h}_1, \hat{h}_2)-\widehat{\mathrm{PHD}}_\mathrm{T}^\ell(\hat{h}_1, h_2^*)\mathrm{|}
    + \mathrm{|}\widehat{\mathrm{PHD}}_\mathrm{T}^\ell(\hat{h}_1, h_2^*) - \mathrm{PHD}_{P_\mathrm{T}}^\ell(h_1^*, h_2^*)\mathrm{|}, \numberthis{\label{ineq:tr1}} \\
    &and \\
    &\leq \mathrm{|}\widehat{\mathrm{PHD}}_\mathrm{T}^\ell(\hat{h}_1, \hat{h}_2)-\widehat{\mathrm{PHD}}_\mathrm{T}^\ell(\hat{h}_2, h_1^*)\mathrm{|}
    + \mathrm{|}\widehat{\mathrm{PHD}}_\mathrm{T}^\ell(\hat{h}_2, h_1^*) - \mathrm{PHD}_{P_\mathrm{T}}^\ell(h_1^*, h_2^*)\mathrm{|} \numberthis{\label{ineq:tr2}}.
\end{align*}
Ineq.~\eqref{ineq:tr2} holds because of the symmetric property of zero-one loss $\ell$, i.e., $\ell(a,b) = \ell(b,a)$. Combining these two inequalities~\eqref{ineq:tr1} and~\eqref{ineq:tr2} with the property of the zero-one loss, $\ell(a,b)-\ell(c,b)=\ell(a,c)$, leads to following inequality:
\begin{align*}
    \mathrm{|}\widehat{\mathrm{PHD}}_\mathrm{T}^\ell(\hat{h}_1, \hat{h}_2) &- \mathrm{PHD}_{P_\mathrm{T}}^\ell(h_1^*, h_2^*)\mathrm{|} \leq \frac{1}{2}\mathrm{|}\widehat{\mathrm{PHD}}_\mathrm{T}^\ell(\hat{h}_1, h_2^*) - \mathrm{PHD}_{P_\mathrm{T}}^\ell(h_1^*, h_2^*)\mathrm{|} \\&+ \frac{1}{2}\mathrm{|}\widehat{\mathrm{PHD}}_\mathrm{T}^\ell(\hat{h}_2, h_1^*) - \mathrm{PHD}_{P_\mathrm{T}}^\ell(h_1^*, h_2^*)\mathrm{|} + \frac{1}{2}(\widehat{\mathrm{PHD}}_\mathrm{T}^\ell( \hat{h}_1, h_1^*)+\widehat{\mathrm{PHD}}_\mathrm{T}^\ell(\hat{h}_2,  h_2^*)).
\end{align*}
To derive the result, we need following inequalities.
\begin{align*}
     \mathrm{|}\mathrm{PHD}_{P_\mathrm{T}}^\ell(\hat{h}_1, h_2^*) &- \mathrm{PHD}_{P_\mathrm{T}}^\ell(h_1^*, h_2^*) \mathrm{|}
    \\ &= \mathrm{|}\mathrm{PHD}_{P_\mathrm{T}}^\ell(\hat{h}_1, h_2^*) - \mathrm{PHD}_{P_\mathrm{T}}^\ell(h_1^*, h_2^*) + 
    \widehat{\mathrm{PHD}}_{\mathrm{T}}^\ell(\hat{h}_1, h_2^*) -
    \widehat{\mathrm{PHD}}_{\mathrm{T}}^\ell(\hat{h}_1, h_2^*)\\ & \hspace{0.3in}+
    \widehat{\mathrm{PHD}}_{\mathrm{T}}^\ell(h_1^*, h_2^*) - \widehat{\mathrm{PHD}}_{\mathrm{T}}^\ell(h_1^*, h_2^*) \mathrm{|},
    \\ &\leq 2\sup\limits_{h \in \mathcal{H}} \mathcal{|}\mathrm{PHD}_{P_\mathrm{T}}^\ell(h, h_2^*) - \widehat{\mathrm{PHD}}_{\mathrm{T}}^\ell(h, h_2^*)\mathcal{|} + 
     \mathrm{|}\widehat{\mathrm{PHD}}_{\mathrm{T}}^\ell(\hat{h}_1, h_2^*)
    - \widehat{\mathrm{PHD}}_{\mathrm{T}}^\ell(h_1^*, h_2^*) \mathrm{|},
    \\&= 2\sup\limits_{h \in \mathcal{H}} \mathcal{|}\mathrm{PHD}_{P_\mathrm{T}}^\ell(h, h_2^*) - \widehat{\mathrm{PHD}}_{\mathrm{T}}^\ell(h, h_2^*)\mathcal{|} + 
    \widehat{\mathrm{PHD}}_{\mathrm{T}}^\ell(\hat{h}_1, h_1^*).
    \numberthis{\label{eq:losspro}}
\end{align*}
Eq.~\ref{eq:losspro} is derived by the property of the zero-one loss, non-negativity and  $\ell(a,b)-\ell(c,b)=\ell(a,c)$.
By Theorem~\ref{thm:bnry_rad}, the following inequality holds with the probability $1-\delta$,
\begin{align*}
     \mathrm{|}\mathrm{PHD}_{P_\mathrm{T}}^\ell(\hat{h}_1, h_2^*) - \mathrm{PHD}_{P_\mathrm{T}}^\ell(h_1^*, h_2^*)\mathrm{|} \leq 
    2\mathfrak{R}_{n_\mathrm{T}}(\mathcal{H}) + 2\sqrt{\frac{\log\frac{4}{\delta}}{2n_\mathrm{T}}} +
    \widehat{\mathrm{PHD}}_{\mathrm{T}}^\ell(\hat{h}_1, h_1^*).
\end{align*}
Again, by Theorem~\ref{thm:bnry_rad}, the following inequality holds with the probability $1-\delta$,
\begin{align*}
    \mathcal{|}\widehat{\mathrm{PHD}}_{\mathrm{T}}^\ell(\hat{h}_1, h_2^*) - \mathrm{PHD}_{P_\mathrm{T}}^\ell(h_1^*, h_2^*)\mathcal{|} \leq 
    3\mathfrak{R}_{n_\mathrm{T}}(\mathcal{H}) + 3\sqrt{\frac{\log\frac{6}{\delta}}{2n_\mathrm{T}}} +
    \widehat{\mathrm{PHD}}_{\mathrm{T}}^\ell(\hat{h}_1, h_1^*).
\end{align*}
Similarly for the second term and by the deifinition of PHD, the following inequality which holds with the probability $1-\delta$:
\begin{equation*}
    \mathrm{|}\widehat{\mathrm{PHD}}_\mathrm{T}^\ell(\hat{h}_1, \hat{h}_2) - \mathrm{PHD}_{P_\mathrm{T}}^\ell(h_1^*, h_2^*)\mathrm{|} \leq 3 \mathfrak{R}_{n_\mathrm{T}}(\mathcal{H}) + 3\sqrt{\frac{\log(\frac{12}{\delta})}{2n_\mathrm{T}}} +
    \widehat{R}_{\mathrm{T}}^\ell(\hat{h}_1, h_1^*) +
    \widehat{R}_{\mathrm{T}}^\ell(\hat{h}_2, h_2^*), \numberthis{\label{ineq:rslt}}
\end{equation*}
which concludes the proof. \qedhere
\end{proof}


\subsection{Proof of Theorem \ref{thm:full}}
\begin{proof}
Combining Theorem~\ref{thm:mainesb} and Theorem~\ref{thm:bnry_rad} concludes the proof.
\end{proof}

\subsection{Proof of Theorem \ref{thmsaito}}
Proof of Theorem \ref{thmsaito} started from the following lemma:
\begin{lemma}\label{lem1}
Let us consider the loss function $\ell$ bounded by a positive constant $M$. Then, for given $h_1$ and $h_2$, and $\delta\in$ (0,1), it holds with probability at least $1-\delta$ that
\begin{align*}
    \mathrm{PHD}_{P_\mathrm{T}}^\ell(h_1, h_2) - \widehat{\mathrm{PHD}}_\mathrm{T}^\ell(h_1,h_2) \leq M\sqrt{\frac{\log{\frac{1}{\delta}}}{2n_\mathrm{T}}}, \\
    \mathcal{|}\mathrm{PHD}_{P_\mathrm{T}}^\ell(h_1, h_2) - \widehat{\mathrm{PHD}}_\mathrm{T}^\ell(h_1,h_2) \mathcal{|} \leq M\sqrt{\frac{\log{\frac{2}{\delta}}}{2n_\mathrm{T}}}
\end{align*}
where $\widehat{\mathrm{PHD}}_\mathrm{T}^\ell(h_1,h_2)=\widehat{R}_\mathrm{T}^\ell(h_1,h_2).$
\begin{proof}
Let two samples $(x_1,\ldots,x_i,\ldots,x_m) \in \mathcal{T}$ and $(x_1,\ldots,x_i^\prime,\ldots,x_m) \in \mathcal{T^\prime}$ be a set of $m \geq 1$  independent random variables and assume that the loss function $\ell$ is bounded by a positive constant M and $h$ and $h^\prime$ are given, then the function $R^\ell(h,h^\prime): \mathcal{X}^m \rightarrow \mathbb{R}$ satisfies that following condition for McDiarmid's inequality \citep{mcdiarmid1989method}:
\begin{equation*}
    \mathcal{|}\widehat{R}_{\mathrm{T}}^\ell(h,h^\prime)-\widehat{R}_{\mathrm{T}^\prime}^\ell(h,h^\prime)\mathcal{|} \leq \frac{M}{m}.
\end{equation*}
Thus McDiarmid's inequality concludes the proof.
\end{proof}
\end{lemma}

\subsection{Bound for multiclass classification in the margin loss}
\begin{theorem}\label{thm3}
Let $\mathcal{H}$ be a set of hypotheses which maps $\mathcal{X}\times\mathcal{Y}$ to $\mathbb{R}$, where $\mathcal{Y}=\{1\ldots,k\}$ and $k$ is the number of output classes. Fix a margin $\rho>0$ and assume that $\ell$ is the zero-one loss and $\ell\rho$ is a margin loss function, $h_1$ and $h_2$ are given. Then, for any $\delta \in$ (0,1) with probability at least $1-\delta$, the following bound holds for all $h\in\mathcal{H}$:
\begin{multline*}
    R_{P_\mathrm{T}}^{\ell}(h,f_\mathrm{T})-R_{P_\mathrm{T}}^{\ell}(h_\mathrm{T}^*,f_\mathrm{T}) \leq \widehat{R}^{\ell_\rho}_{\mathrm{T}}(h, h_1) + \widehat{\mathrm{PHD}}_\mathrm{T}^{\ell}(h_1, h_2) + R_{P_\mathrm{T}}^{\ell}(h_2, h_\mathrm{T}^*) + \frac{4k}{\rho}\mathfrak{R}_{n_\mathrm{T}}(\Pi_1(\mathcal{H})) + 2\sqrt{\frac{\log{\frac{2}{\delta}}}{2n_\mathrm{T}}},
\end{multline*}
where $\widehat{R}^{\ell_\rho}_{\mathrm{T}}(h, h^\prime)=\mathbb{E}_{\mathrm{T}}[1_{h(x,y)-\max_{y^\prime\ne y}h(x,y^\prime)\leq \rho}]$ for $y=\argmax\limits_{y\in\mathcal{Y}} h^\prime(x,y)$ and $\Pi_1(\mathcal{H}) = \{(x, y) \rightarrow h(x,y) : y \in \mathcal{Y}$, $h \in \mathcal{H}\}$.

\begin{proof}
Combining Theorem~\ref{thm:mlt_rad} and Lemma~\ref{lem1} concludes the proof.
\end{proof}
\end{theorem}

\section{Datasets and Settings}
To evaluate method more precisely, we divide datasets into disjoint subsets. 

\subsection{Datasets used in Section \ref{subsec_cmpr}}
\noindent \textbf{MNIST\citep{lecun1998mnist}:} We divide train data into two parts, 50,000 for source train and 10,000 for source test, and use MNIST test data as target train and test data. And label them as odd/even.

\noindent \textbf{EMNIST\citep{cohen2017emnist}:} extended MNIST dataset which has 4 times more data than MNIST. We divide train data into three parts, 40,000 for source train, 160,000 for target train, 40,000 for source test, and use EMNIST test for target test. And label them as odd/even.

\subsection{Datasets used in Section \ref{sec:exp_infeasible}}

\noindent \textbf{MNIST, MNISTM\citep{ganin2016domain}:} 10,000 for source train, source and target test, 30,000 for target train.

\noindent \textbf{EMNIST:} 40,000 for source train, 160,000 for target train, 20,000 for both
    source and target train.
    
\noindent \textbf{CIFAR-10\citep{krizhevsky2014cifar}, SVHN\citep{netzer2011reading}:} 10,000 for source train, 5,000 for source and target test, 30,000 for target train.

\subsection{Architecture of networks}
For simplicity, we denote convolutional neural network as Conv[out\_channel, kernel size, stride, zero-padding], max pooling layer as Max[kernel size, stride] and average pooling layer as Avg[kernel size, stride]. []*$n$ means there are $n$ such layers.
\subsubsection{MLP in PNU}~\label{sec:mlppnu}
five-layer fully connected neural network with rectifier (ReLU) \citep{nair2010rectified} as activation function: d-512-512-512-512-1
\subsubsection{MLP in VAT}~\label{sec:mlpvat}
five-layer fully connected neural network with Leaky rectifier (LReLU) \citep{maas2013rectifier} which slope is 0.1 as activation functions: d-512-512-512-512-1. Batch normalization layer \citep{ioffe2015batch} was applied before hidden layers.
\subsubsection{CNN in VAT}
Conv[128, 3, 1, 1]*3 - Max[2, 2] - Dropout -
Conv[256, 3, 1, 1]*3 - Max[2, 2] - Dropout -
Conv[512, 3, 1, 0] - Conv[256, 1, 1, 0] - Conv[128, 1, 1, 0] - Avg[2, 2] - Full Connected layer.

Note that after every convolution layer we use batch normalization and leaky rectifier as activation functions. We changed the input image size as [32, 32] if image size is different from [32, 32].

\subsubsection{MLP in Section~\ref{sec:source_selection}}
\noindent \textbf{PHD and Classifier in the target domain:} same architecture with MLP in VAT in Section~\ref{sec:mlpvat}. \\
\noindent \textbf{Wasserstein-1 distance:} same architecture with MLP in VAT in Section~\ref{sec:mlppnu}. \\


\section{Experiments results}
\subsection{binary setting}

\begin{table}[H]
\caption{Empirical generalization error bound of PHD in the binary setting, MLP model}
\begin{center}
\begin{tabular}{cccccc}
        \toprule
        &&&&\multicolumn{2}{c}{Infeasible Term} \\
         Source & Target & $h_\mathrm{S}$(\%) & $\mathrm{PHD}_\mathrm{PNU}$ & $R_{P_\mathrm{T}}^{\ell_{01}}(h_\mathrm{PNU}, h_\mathrm{T}^*)$ & $R_{P_\mathrm{T}}^{\ell_{01}}(h_\mathrm{S}, h_\mathrm{T}^*)$   \\
         \midrule
         MNIST & MNIST  & 97.828 & 0.0071 (0.001) & 0.0176 (0.001) & \textbf{0.0172 (0.001)} \\
         MNIST & CIFAR-10 & 97.828 & 0.6828 (0.229) & 0.4874 (0.043) & 0.4951 (0.060) \\
        \bottomrule
\end{tabular}
\end{center}
\end{table}
Differently from the case where two domains are identical, PHD and two infeasible terms give a high value when two domains are not related. This is very intuitive since these values represent the distance between two domains. Note that value around 0.5 means random guess in the binary setting, which shows that two hypotheses are not related in the target domain.

\subsection{multiclass setting}
\begin{table}[H]
\caption{Empirical  generalization error bound of PHD in the multiclass setting}
\begin{center}
\begin{tabular}{ccccc|cc}
     \toprule
     &&&&&\multicolumn{2}{c}{Infeasible Term} \\
       Source & Target & Model & $h_\mathrm{S}$(\%) & $\mathrm{PHD}_\mathrm{VAT}$ & $R_{P_\mathrm{T}}^{\ell_{01}}(h_\mathrm{VAT}, h_\mathrm{T}^*)$ & $R_{P_\mathrm{T}}^{\ell_{01}}(h_\mathrm{S}, h_\mathrm{T}^*)$  \\
    \midrule
       \multirow{2}{*}{MNISTM} & \multirow{2}{*}{MNISTM}& VGG11 & 96.48 (0.169) & 0.0349 (0.002) & \textbf{0.0294 (0.002)} & 0.0333 (0.002) \\
      && CNN & 98.03 (0.117) & 0.0189 (0.002) & \textbf{0.0166 (0.001)} & 0.0208 (0.002) \\   
      \midrule  
      \multirow{3}{*}{SVHN} &\multirow{3}{*}{SVHN}& MLP &  75.73 (0.601) & 0.1833 (0.006) & \textbf{0.2246 (0.006)} & 0.2374 (0.006) \\
      && VGG11 & 90.39 (0.339) & 0.0852 (0.005) & \textbf{0.0817 (0.006)} & 0.0898 (0.003) \\ 
      && CNN & 91.83 (0.407) & 0.0729 (0.005) & \textbf{0.0652 (0.003)} & 0.0751 (0.004) \\ 
     \midrule
      \multirow{2}{*}{CIFAR-10} & \multirow{2}{*}{CIFAR-10} & VGG11 & 73.66 (0.817) & 0.2223 (0.008) & \textbf{0.2348 (0.013)} & 0.2415 (0.010) \\
      &&CNN & 71.21 (0.598) & 0.2012 (0.008) & \textbf{0.2118 (0.006)} & 0.2184 (0.008) \\
      \midrule
      \multirow{1}{*}{MNIST} & \multirow{1}{*}{SVHN}& MLP & 96.72 (0.329) & 0.8888 (0.040) & 0.8852 (0.0301) & 0.8695 (0.005) \\
    \midrule
      \multirow{4}{*}{MNISTM} &\multirow{2}{*}{SVHN}& VGG11 & 95.18 (0.384) & 0.8589 (0.072) & 0.8693 (0.045) & 0.6391 (0.008) \\ 
      && CNN & 96.69 (0.280) & 0.8647 (0.045) & 0.8980 (0.014) & 0.7614 (0.011) \\ 
      &\multirow{2}{*}{CIFAR-10}& VGG11 & 95.03 (0.349) & 0.8960 (0.022) & 0.9007 (0.005) & 0.9223 (0.006) \\
      && CNN & 96.63 (0.484) & 0.8142 (0.043) & 0.9145 (0.023) & 0.9167 (0.005) \\ 
      \bottomrule
\end{tabular}
\end{center}
\end{table}
As the model becomes more complex, our proposed method PHD can be estimated more accurately differently from existing methods even in the multiclass setting. When domains are not identical, PHD and two infeasible terms are supposed to be high, and estimation of it also gives high values which support informativeness of our proposed discrepancy measure. Note that value around 0.9 means random guess which may imply two domains are not related. 

\subsection{Wasserstein-1 distance}
As a baseline, the Wasserstein-1 distance, which does not utilize any source label information, was calculated. We used a architecture following \citet{radford2015unsupervised} with gradient penalty \citep{arjovsky2017wasserstein}.
\begin{table}[H]
\caption{Wasserstein-1 distance as a baseline measure.}
\begin{center}
\begin{tabular}{lccccccc}
        \toprule
         Source & MNIST & SVHN & MNISTM & CIFAR-10 & MNIST & MNISTM & MNISTM \\ \\
         Target & MNIST & SVHN & MNISTM & CIFAR-10 & SVHN & CIFAR-10 & SVHN \\
         \midrule
         & 0.3893 & 0.1465 & 0.1642 & 0.2008 & 17.14 & 7.305 & 8.036 \\
         \bottomrule
\end{tabular}
\end{center}
\end{table}

\subsection{Source selection task}
\begin{table}[H]
\caption{Source selection task}
\label{tab:src}
\begin{center}
\resizebox{\textwidth}{!}{\begin{tabular}{cc|ccccccccc}
     \toprule
     &&\multicolumn{9}{c}{Noise rate $(\sigma)$} \\
       Evaluation & Methods & 0.1& 0.15&0.2&0.25&0.3&0.35&0.4&0.45&0.5\\
    \midrule
       \multirow{2}{*}{ACC} & $\mathrm{PHD}$ & 93.55 (0.231) & 93.36 (0.364) & 93.57 (0.418) & 93.47 (0.364)
       & 93.31 (0.429) & 93.4 (0.431) & 93.49 (0.667) & 93.457 (0.530) & 93.53 (0.526)\\
      & W1 & 93.12 (0.260) & 93.35 (0.264) & 93.28 (0.364) & 92.74 (0.513) & 92.68 (0.609) & 91.98 (1.034)
      & 91.013 (1.354) & 91.23 (1.115) & 90.37 (1.687) \\
      \midrule 
      \multirow{2}{*}{Score} & $\mathrm{PHD}$ & 2.5 (0.67) & 2.5 (0.92) & 3.4 (0.66) & 3.6 (0.49) & 3.6 (0.49) & 3.9 (0.70) &4.1 (0.7) & 3.8 (0.4) & 4.4 (0.66)\\   
      & W1 & 2.3 (0.64) & 2.1 (1.04) & 2.2 (0.6) & 2.1 (0.54) & 1.8 (0.75) & 1.7 (0.9) & 1.5 (0.67) & 1.8 (0.6) & 1.5 (0.81)\\
      \bottomrule
\end{tabular}}
\end{center}
\end{table}
From Table~\ref{tab:src}, we can see that our proposed method $\mathrm{PHD}$ maintained the accuracy in the target domain by selecting clean sources successfully compared to the Wasserstein-1 distance (W1) which failed to select clean sources and degraded the overall performance in the target domain.
Even though the degradation is not significant, it can be interpreted as our proposed method can prevent even small degradation in the target domain by appropriately estimating the discrepancy between domains, which might be overlooked by the W1 distance.
\end{document}